%% file: flexicache.tex
\documentclass{article}

\usepackage{microtype}
\usepackage{graphicx}
\usepackage{subfigure}
\usepackage{amsmath}
\usepackage{booktabs} 
\usepackage{xcolor}
\usepackage{amssymb}
\usepackage{enumitem}
\usepackage{makecell}
\usepackage{booktabs}
\usepackage{multirow}
\usepackage{hyperref}
\usepackage{listings}

\lstdefinestyle{bashstyle}{
    backgroundcolor=\color{gray!10},
    frame=single,
    basicstyle=\ttfamily\small,
    keywordstyle=\color{blue},
    commentstyle=\color{green!50!black},
    stringstyle=\color{orange},
    breaklines=true,
    columns=fullflexible,
    keepspaces=true
}

\usepackage{hyperref}

\newcommand{\parab}[1]{\vspace{0.01in}\noindent\textbf{#1}}

\usepackage[accepted]{mlsys2026}

\newcommand{\rev}[1]{\textcolor{black}{#1}}

\mlsystitlerunning{FlexiCache: Leveraging Temporal Stability of Attention Heads for Efficient KV Cache Management}

\AddToShipoutPictureBG*{
  \AtPageUpperLeft{
    \hspace*{\paperwidth}
    \raisebox{-68pt}{
      \llap{
        \href{https://www.acm.org/publications/policies/artifact-review-and-badging-current}{
          \includegraphics[height=65pt]{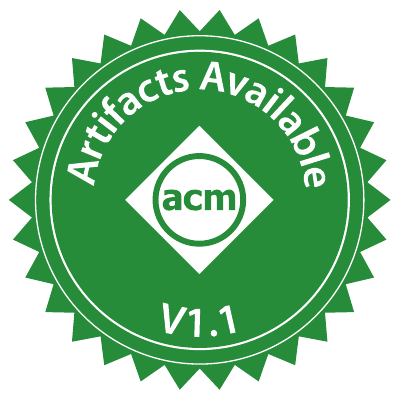}}
        \hspace{1pt}
        \href{https://www.acm.org/publications/policies/artifact-review-and-badging-current}{
          \includegraphics[height=65pt]{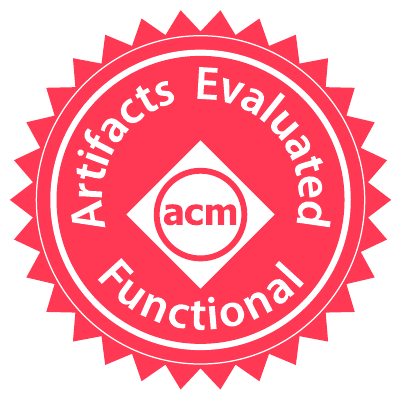}}
        \hspace{1pt}
        \href{https://www.acm.org/publications/policies/artifact-review-and-badging-current}{
          \includegraphics[height=65pt]{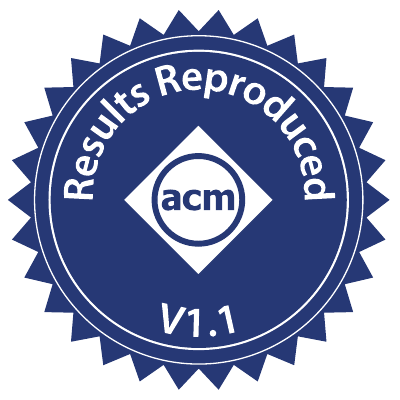}}
        \hspace{80pt}
      }
    }
  }
}

\begin{document}

\twocolumn[
\mlsystitle{FlexiCache: Leveraging Temporal Stability of Attention Heads for Efficient KV Cache Management}

\mlsyssetsymbol{equal}{*}

\begin{mlsysauthorlist}
\mlsysauthor{Nazmul Takbir}{uci}
\mlsysauthor{Hamidreza Alikhani}{uci}
\mlsysauthor{Nikil Dutt}{uci}
\mlsysauthor{Sangeetha Abdu Jyothi}{uci}
\end{mlsysauthorlist}

\mlsysaffiliation{uci}{Department of Computer Science, University of California, Irvine}

\mlsyscorrespondingauthor{Nazmul Takbir}{ntakbir@uci.edu}

\mlsyskeywords{Efficient LLM Inference, Long-context LLM, KV cache offloading, Sparse Attention, Attention head stability}

\vskip 0.3in

\begin{abstract}
Large Language Model (LLM) serving is increasingly constrained by the growing size of the key-value (KV) cache, which scales with both context length and generation length. Prior work shows that attention is dominated by a small subset of critical tokens, yet existing systems struggle to exploit this efficiently without degrading accuracy, especially in long generation. \rev{We make a key observation: the temporal stability of these critical tokens varies significantly across KV heads. Some heads consistently focus on the same tokens, while others shift frequently.} Building on this insight, we introduce FlexiCache, a hierarchical KV-cache management system that leverages the temporal stability of KV heads to reduce GPU memory usage and computation overhead, while preserving model accuracy. FlexiCache classifies KV heads as stable or unstable: it retains all KV-cache pages from unstable heads in GPU memory, whereas for stable heads, it keeps only the top-K pages on the GPU and offloads the rest to host memory. By exploiting temporal stability, FlexiCache performs periodic reranking for stable heads to fetch newly promoted top pages. Implemented atop vLLM, FlexiCache reduces GPU memory footprint for long-context requests by up to \textbf{70\%}, improves offline serving throughput by \textbf{1.38–1.55×}, and lowers online token latency by \textbf{1.6–2.1×}, all while maintaining accuracy in long-context, long-generation scenarios.
\end{abstract}
]



\printAffiliationsAndNotice{}  

\input{sections/intro}

\input{sections/motivation}

\input{sections/system}

\input{sections/eval}

\input{sections/discussion_future}

\input{sections/related}

\input{sections/conclusion}

\bibliography{flexicache}
\bibliographystyle{mlsys2026}

\clearpage

\input{sections/artifact_appendix}

\end{document}

%% file: sections/intro.tex
\section{Introduction}
\label{introduction}
The rapid adoption of Large Language Models (LLMs) has created an urgent need for efficient and scalable serving systems. As use cases grow in complexity, LLMs continue to improve at processing longer contexts and generating longer outputs. Context lengths of 32k–128k tokens are standard in open-source models~\cite{dubey2024llama, yang2025qwen3, jiang2024mixtral}, with the latest state-of-the-art systems exceeding million tokens~\cite{comanici2025gemini, yang2025qwen2}. Meanwhile, tasks such as large-scale code synthesis and long-form research writing have pushed generation lengths into the range of several thousand tokens, with recent benchmarks evaluating LLMs on outputs as long as 32k tokens~\cite{wu2025longgenbench, bai2025longwriter}.

Serving requests with long contexts introduces significant systems challenges, primarily due to the large size of the key–value (KV) cache. These challenges are further exacerbated when long contexts are combined with long generations: not only is each request’s KV cache large, but it must also remain resident in GPU memory for extended periods and be repeatedly accessed at every decoding step. Since the decode phase is memory-bound, a common strategy to improve GPU utilization is to increase batch size \cite{kwon2023efficient}. However, large KV caches quickly exhaust GPU memory, limiting the achievable batch size and reducing overall throughput.

Prior work has observed that, at each decoding step, a small subset of tokens dominates the attention output. StreamingLLM \cite{xiao2023efficient} retains a sliding window of recent tokens and the first few tokens, but this static, request-independent strategy degrades accuracy on long-context tasks. SnapKV \cite{li2024snapkv} and MorphKV \cite{ghadia2025dialogue} improve on this by using attention scores to identify and discard the KV cache of less important tokens in a request-dependent manner. While effective for long contexts, they underperform in long generations because discarded KV cannot be reused even if it later becomes important. Quest \cite{10.5555/3692070.3694025} mitigates this by re-selecting top-K KV pages at each step through query-aware scoring, reducing attention computation and I/O but not GPU memory use. Finally, LServe \cite{yang2025lserve} leverages DuoAttention \cite{xiao2025duoattention} to convert some heads into streaming heads whose attention is mostly local or $\Lambda$-shaped, discarding their KV pages while keeping full KV for others. This achieves a better balance between efficiency and accuracy, but permanently discarding large portions of the KV cache for half of the heads can degrade accuracy on long-context, long-generation workloads.

We make a key observation: KV heads exhibit varying temporal stability in their set of top-K KV pages across decode steps. We analyze the overlap of top-K KV page sets between consecutive decode steps for each head independently and observe a distinct pattern: some heads (\textit{stable} heads) exhibit high overlap that decays slowly, while others (\textit{unstable} heads) show consistently low overlap. Importantly, these stability patterns are model-intrinsic: a head that is stable in one task remains stable across other tasks.

Leveraging the temporal stability of KV heads, we present FlexiCache, a hierarchical KV-cache management system for efficient LLM serving. FlexiCache applies sparse attention over the top-K KV pages across all heads to reduce the computational cost of attention during decoding. To minimize GPU memory usage, FlexiCache first performs a one-time offline classification of KV heads as \textit{stable} or \textit{unstable}. During serving, it retains only the top-K pages from stable heads in GPU memory and offloads the rest to host memory. FlexiCache periodically re-ranks the KV pages of stable heads, fetching only newly promoted top-K pages from host memory, thereby reducing I/O overhead. For unstable heads, whose top-K sets change frequently, all KV pages remain in GPU memory to avoid repeated host–GPU transfers. By combining sparse attention with head-level temporal stability, FlexiCache effectively reduces computation, GPU memory footprint, and host–GPU data movement while maintaining high language performance.

We implement FlexiCache atop vLLM and evaluate it across a broad suite of long-context, long-generation language modeling tasks. We also assess system performance under diverse scenarios, including offline serving, online serving, and microbenchmarks targeting key optimizations. FlexiCache achieves 99\% of the baseline model’s accuracy across tasks while reducing the GPU memory footprint by up to 70\% for long-context requests and accelerating the decode kernel by up to 4× in large batched decode settings. These improvements translate into 1.38–1.55× higher end-to-end throughput in offline serving and 1.6–2.1× lower token latency in online serving.

In summary, we make the following contributions:
\begin{itemize}[leftmargin=*,nolistsep]
    \item We identify that KV heads exhibit model-intrinsic temporal stability in selecting their top KV pages for attention computation.
    \item We develop FlexiCache, an LLM serving system with hierarchical KV-cache management that leverages the temporal stability of KV heads.
    \item We show that FlexiCache, built atop vLLM, reduces GPU memory usage by up to 70\%, increases offline serving throughput by up to 1.55$\times$, reduces token latency in online serving by up to 2.1$\times$, and preserves model accuracy in long-context, long-generation settings.
    \item \rev{The FlexiCache repository is available as open-source at: \href{https://github.com/NazmulTakbir/FlexiCache}{https://github.com/NazmulTakbir/FlexiCache}}
\end{itemize}

%% file: sections/motivation.tex
\section{Background \& Motivation}
LLMs based on the transformer architecture \cite{vaswani2017attention} perform inference in two stages: prefill and decode. In the prefill stage, all input tokens are processed in parallel, and the model produces query (Q), key (K), and value (V) vectors for each token. The K and V vectors are stored in the KV cache for later use. In the decode stage, the model generates one token at a time, appending a new K/V pair at each step. The decode stage is memory-bound, and a common strategy to increase decode throughput is to use a higher batch size. However, batch size is ultimately constrained by the GPU memory available to hold all active requests’ KV caches. Moreover, GPU memory fragmentation further limits the effective batch size. PagedAttention \cite{kwon2023efficient} addresses this by storing KV in fixed-size pages dereferenced via a block table, which minimizes fragmentation and enables larger batch sizes. 

Since the KV cache grows linearly with context length, a large context length not only limits batch size but also increases the compute and I/O overhead of self-attention, as each decode step fetches the entire KV cache. Prior work has shown that, at each decoding step, a small subset of KV cache pages dominates attention. Quest~\cite{10.5555/3692070.3694025} introduced a method to identify these subsets independently for each attention head at each decoding step. While selecting new top-K pages at every step preserves language-modeling accuracy, it incurs substantial computational overhead and requires the full KV cache to remain resident on the GPU. To alleviate this cost, LServe\cite{yang2025lserve} reranked only half of the heads, updating them periodically (every four decoding steps). However, this partial reranking strategy can lead to accuracy degradation, particularly in long-generation scenarios, since heads that are not reranked may permanently discard KV entries that later become relevant.

\subsection{Temporal Stability in Top-K Selection}
We make the key observation that many KV heads exhibit temporal stability in their top-K page selections: the same pages tend to remain among the most attended across consecutive decoding steps. Moreover, within a given model, the degree of stability varies systematically across heads—some remain highly stable, whereas others are persistently unstable. In the following analysis, we (i) formalize a quantitative measure of this temporal stability, (ii) identify the least stable heads for each model and demonstrate that their instability pattern is model-intrinsic and consistent across tasks, and (iii) leverage these insights to design an efficient KV-cache management mechanism.

\begin{figure}[t]
  \centering
  \includegraphics[width=\columnwidth]{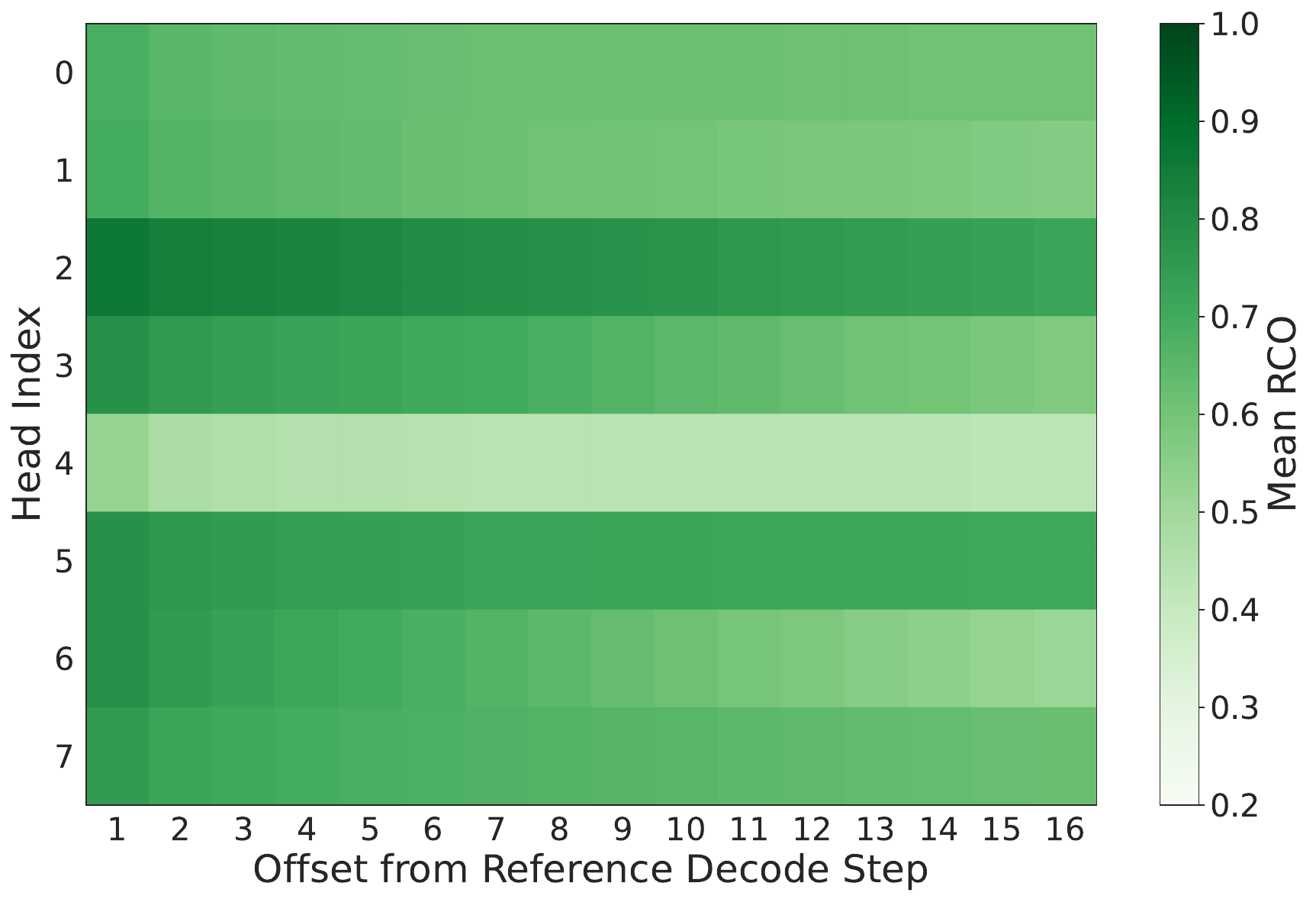}
  \vspace{-5mm}
  \caption{\textbf{Temporal stability patterns of KV heads}. In the Llama-3.1-8B-Instruct layer 4, some heads maintain high RCO across offsets, while others show persistently low values.}
  \label{fig:overlap-demo}
\end{figure}

\subsection{Quantifying Temporal Stability of KV Heads}
\label{subsec:temporal-stability}
We begin by collecting the set of top-\(K\) page indices at each decode step for samples from several long-context, long-generation tasks in LongBench \cite{bai-etal-2024-longbench} and L-Eval \cite{an-etal-2024-l}. Concretely, for a given sample, we store the data in a tensor of shape \([D, L, H, K]\), where \(D\) denotes the number of decoding steps, \(L\) the layer index, \(H\) the KV-head index within the layer, and \(K\) the number of selected top-\(K\) pages. In other words, for each sample and each layer–head pair, we record the indices of the top-\(K\) pages attended to at every decoding step.

\rev{For layer \(\ell\), head \(h\), and decoding step \(s\), let \(S_{\ell,h}(s)\) denote the set of top-\(K\) page indices for a particular sample. We compute the degree of overlap between \(S_{\ell,h}(s)\) and the subsequent top-\(K\) selections of the same head in the same layer within a fixed window of \(W\) steps. We set the stability window size W to match the stable-head reranking interval (see section \ref{sec:kv_page_scoring}).} To isolate genuine temporal persistence from overlap due to random chance, we compute the \emph{random-corrected overlap} (RCO) between the page selections at step \(s\) and step \(t = s + \Delta\) (\(1 \le \Delta < W\)) as:
\rev{
\[
\mathrm{RCO}_{\ell,h}(s,t) =
\max\!\left(
0,\,
\frac{
\lvert S_{\ell,h}^{(s)} \cap S_{\ell,h}^{(t)} \rvert / K - K / N_t
}{
1 - K / N_t
}
\right)
\]
}
where \(N_t\) is the number of candidate pages available at step \(t\). This formulation subtracts the expected overlap under a hypergeometric random-draw model (\(K / N_t\)) and normalizes by the maximum value \(1 - K / N_t\), which represents the gap between random-chance overlap
and perfect overlap. Intuitively, the RCO quantifies how much better the observed overlap is than random chance. An  $\mathrm{RCO}$ of 0 indicates random-level overlap, and 1 indicates perfect overlap.

We aggregate RCO values across all pairs in a window and define a \emph{temporal stability score} for each KV head as:
\rev{
\[
TS_{\ell,h}(s) = \frac{1}{W - 1} \sum_{t = s + 1}^{s + W - 1} \mathrm{RCO}_{\ell,h}(s, t)
\]
}
Lower stability values indicate greater variation in a head’s attention focus across decoding steps within the window. Figure~\ref{fig:overlap-demo} illustrates this behavior for the eight KV heads of layer~4 in Llama-3.1-8B-Instruct on the \textit{SPACE} (hotel review summarization) task from L-Eval. For each offset from the reference decoding step, we plot the mean RCO averaged across all windows and samples. Some heads consistently exhibit low RCO (unstable), while others maintain high RCO that gradually decays with offset (stable).

\begin{table}[t]
\centering
\footnotesize
\setlength{\tabcolsep}{3pt}
\caption{\textbf{Cross-task overlap of unstable heads}. The values are the intersection size normalized by the unstable-head set size (\(|A_i \cap A_j|/64\)), in Llama-3.1-8B-Instruct. GovReport is from LongBench \cite{bai-etal-2024-longbench}; the rest from L-Eval \cite{an-etal-2024-l}.}
\label{tab:cross-task-overlap}
\resizebox{\columnwidth}{!}{%
\begin{tabular}{lcccccccc}
\toprule
\textbf{Dataset} &
\makecell[c]{Open-\\review} &
\makecell[c]{Big-\\Patent} &
\makecell[c]{Multi-\\News} &
\makecell[c]{QM-\\Sum} &
\makecell[c]{Gov-\\Report} &
\makecell[c]{SP-\\ACE} &
\makecell[c]{CU-\\AD} &
\makecell[c]{Summ-\\Screen} \\
\midrule
Openreview   & 1.00 & 0.81 & 0.78 & 0.89 & 0.92 & 0.83 & 0.73 & 0.81 \\
BigPatent    &   0.81   & 1.00 & 0.72 & 0.86 & 0.83 & 0.75 & 0.77 & 0.77 \\
Multi-News   &   0.78   &   0.72   & 1.00 & 0.80 & 0.80 & 0.77 & 0.78 & 0.80 \\
QMSum        &   0.89   &   0.86   &  0.80    & 1.00 & 0.92 & 0.84 & 0.75 & 0.86 \\
GovReport    &   0.92   &   0.83   &  0.80    & 0.92     & 1.00 & 0.84 & 0.80 & 0.83 \\
SPACE        &   0.83   &   0.75   &    0.77   & 0.84     &   0.84   & 1.00 & 0.70 & 0.91 \\
CUAD         &  0.73    &   0.77   &   0.78   & 0.75     &   0.80   &   0.70   & 1.00 & 0.67 \\
SummScreen   &  0.81   &   0.77    &   0.80   & 0.86     &   0.83   &   0.91   &   0.67   & 1.00 \\
\bottomrule
\end{tabular}%
}
\end{table}
\begin{figure*}[t]
  \centering
  \includegraphics[width=0.67\textwidth]{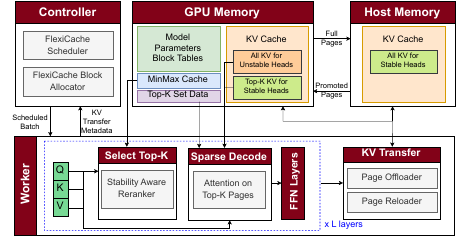}
  \caption{\textbf{FlexiCache system architecture}. At the worker, the top-K selector identifies the most relevant KV pages for each head and updates them at different frequencies based on head stability. The sparse decode kernel attends only to these selected pages. GPU memory stores the full KV cache of unstable heads and only the top-K pages of stable heads, with the rest in host memory. The block allocator manages this hierarchical KV layout, while the KV transfer module and scheduler pipeline host–GPU KV transfers with computation.}
  \label{fig:flexicache-system-A}
\end{figure*}
\subsection{Identifying Model-Intrinsic Stability Patterns}
Given the per-window temporal stability scores for every KV head, we classify the heads into two categories---stable and unstable---by selecting the least stable 25\% of heads as unstable and the remaining 75\% as stable. Concretely, across all decoding windows within and across all samples, we sort the KV heads by their stability scores and designate the most frequent heads in the bottom quartile as \textit{unstable}, while the remainder are considered \textit{stable}.

We perform the procedure independently on each evaluation task, resulting in one set of unstable heads per task for a given model. Each of these sets has the same cardinality (25\% of the total heads), enabling direct comparison across tasks. To assess whether head instability is an intrinsic property of the model, we measure pairwise overlap between the unstable head sets across different tasks for the same model.
The resulting cross-task overlap matrix for Llama-3.1-8B-Instruct is presented in Table \ref{tab:cross-task-overlap}. The overlaps are consistently high, with a mean overlap of 0.83 across tasks and several pairs reaching close to 0.90, demonstrating that the identity of unstable heads is largely preserved across tasks. Thus, head instability is a model-intrinsic characteristic: certain heads inherently exhibit volatile KV-page selection behavior across decoding steps. Furthermore, because unstable heads are consistent across tasks, a single offline profiling \footnote{\rev{Profiling on GovReport takes $\sim$2 hours for Llama-3.1-8B, dominated by per-step, per-head logging of top-K page indices. As a one-time offline step per model, it is not on the critical path of online serving.}} run per model suffices to identify them for subsequent online serving. We observe similarly high cross-task overlaps for other models (Mistral-7B, Mistral-Small-24B, and Qwen2.5-32B); \rev{detailed in section \ref{sec:model-instrinsic-unstable-heads}.}

\rev{While stability scores of heads form a continuous spectrum, we adopt a binary classification (stable vs. unstable) for system simplicity. In principle, a finer-grained categorization (e.g., multiple stability tiers with different reranking intervals or top-K budgets) could further specialize KV management policies. However, such designs introduce additional control complexity in block-table management, scheduling, and KV transfers. By restricting the design to two classes, we keep the control-plane overhead minimal.}

%% file: sections/system.tex
\section{Flexicache}
\label{flexicache}
We present FlexiCache, an LLM-serving system that leverages the inherent sparsity and temporal stability of KV heads to manage KV cache efficiently. The key goals of FlexiCache are (i) \textbf{\textit{G1}}: reduce attention computation overhead, (ii) \textbf{\textit{G2}}: optimize GPU memory usage, (iii) \textbf{\textit{G3}}: minimize I/O transfers, and (iv) \textbf{\textit{G4}}: maintain high language modelling quality. FlexiCache achieves these goals simultaneously, even under long context and long generation scenarios, by prioritizing and retaining only the most critical KV cache pages in GPU memory.

\subsection{FlexiCache System Overview}
FlexiCache employs a multi-pronged design to meet its objectives. Leveraging the temporal stability of critical KV-page selection, it first classifies the heads as \textit{stable} or \textit{unstable}. To reduce computational load (\textit{G1}), FlexiCache employs sparse attention, using only the top-K KV pages for attention computation across all heads during the decode phase. To improve GPU memory efficiency (\textit{G2}), it keeps only the top-K KV pages from stable heads in GPU memory, while retaining all KV pages from unstable heads on the GPU. The remaining KV pages of stable heads are stored in host memory. To minimize host–GPU data transfer (\textit{G3}), FlexiCache controls the frequency of re-ranking: unstable heads are re-ranked at every step, but since all their KV pages reside on the GPU, no transfers are needed; stable heads are re-ranked only periodically, and after each re-ranking, only the newly promoted top-K pages are fetched from host memory. Finally, by jointly exploiting KV-head stability and attention sparsity, FlexiCache ensures that the top-K page selection operates over the entire KV cache (since no pages are permanently discarded), thereby sustaining high language modeling performance (\textit{G4}).

FlexiCache is implemented on top of vLLM, and its main components are illustrated in Figure~\ref{fig:flexicache-system-A}. The stability-aware re-ranker identifies the top-K KV pages for each head independently and at different frequencies based on their stability. The sparse decode module performs attention computation using only the top-K KV pages for each head. The KV transfer module manages data movement between GPU and host memory: it offloads all KV pages of stable heads to host memory and periodically reloads only the newly promoted top-K pages after each re-ranking.

The FlexiCache controller orchestrates request execution at the workers and includes two key components. The FlexiCache Scheduler efficiently pipelines computation and KV reloads across batched requests to maximize GPU utilization. The FlexiCache Block Allocator extends vLLM’s allocator with hierarchical memory awareness, allocating just enough GPU KV blocks for the top-K KV pages of stable heads and for the entire KV cache of unstable heads, while assigning host KV blocks to store the full KV cache of stable heads.

\subsection{KV-Page Scoring Mechanism}
\label{sec:kv_page_scoring}
FlexiCache uses a page importance scoring mechanism similar to that in Quest \cite{10.5555/3692070.3694025} and LServe \cite{yang2025lserve}. For each KV page, we maintain two vectors---the per-dimension minimum and maximum key values within that page. 
Given the current query vector $q$, the importance of a page $p$ is estimated as the maximum possible dot product between $q$ and any key in that page:
\[
s_p = \sum_i \max(q_i \cdot k^{\min}_{p,i},\; q_i \cdot k^{\max}_{p,i})
\]
However, FlexiCache differs in two key ways. First, page scoring operates at different frequencies for stable and unstable heads: unstable heads are scored at every step, \rev{whereas stable heads are scored periodically (every 16 steps). Owing to their temporal stability, stable heads reuse the most recently computed top-K page set between reranking steps.} If a layer consists solely of stable heads, scoring for that layer can often be skipped entirely. Second, FlexiCache decouples the min–max vectors from the KV cache pages by storing them in a dedicated MinMax cache that remains on the GPU even when the corresponding KV pages are offloaded to host memory. This enables the system to evaluate the importance of an offloaded page and reload it to GPU memory when necessary. The MinMax cache maintains its own block manager and allocates large blocks, each containing min–max vectors for up to 128 KV cache blocks, to minimize allocation overhead. Together, these optimizations substantially reduce the overhead of page scoring and ranking.

\begin{figure}[t]
  \centering
  \includegraphics[width=0.75\columnwidth]{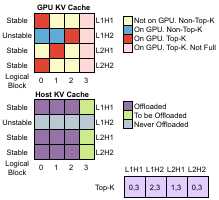}
  \caption{\textbf{Hierarchical KV-cache placement}. The diagram illustrates the mapping of a request with four logical KV pages running on a two-layer, two-head model.}
  \label{fig:kv-placement-example}
\end{figure}

\subsection{Hierarchical KV Cache Management}
The hierarchical management of the KV cache builds on the temporal stability of KV heads (\S~\ref{subsec:temporal-stability}) in selecting their top-K KV pages. We classify the least stable 25\% of heads as unstable and the remaining heads as stable. For stable heads, FlexiCache keeps only the top-K most important KV pages in GPU memory, while offloading the rest to host memory. In contrast, for unstable heads, it retains the entire KV cache on the GPU.

Figure~\ref{fig:kv-placement-example} illustrates this hierarchical placement using a toy example of a model with two layers and two KV heads per layer. The request contains four logical pages, three of which are full, with a top-K size of two. For stable heads, only the two top-K pages reside in GPU memory, while the host memory stores all three full KV pages. The last partially filled page, which is always in the top-K since it is being appended to, will be offloaded to host memory once it becomes full. For the unstable head, all four KV pages remain in GPU memory and are never offloaded.

To make hierarchical KV-cache management practical, FlexiCache minimizes the overhead of moving KV pages between host and GPU memory in the following ways:

\parab{Minimizing Transfer Size.}
\label{sec:minimize-tx-size}
In the host-to-GPU direction, we transfer only the newly promoted pages---those that were not in the previous top-K set but appear in the current one---after the periodic reranking of top-K pages for stable heads. This reduces the frequency and size of transfers, as the top-K set itself represents only a fraction of the total KV cache, while temporal stability further reduces the difference between top-K sets. To minimize GPU-to-host transfers, each KV page of a stable head is offloaded exactly once. A large offload occurs after the prefill phase, followed by small incremental offloads as newly filled pages are completed during decoding. Consequently, host memory always holds the full KV cache for stable heads, while the GPU retains only the top-K pages. 
\begin{figure}[t]
  \centering
  \includegraphics[width=0.9\columnwidth]{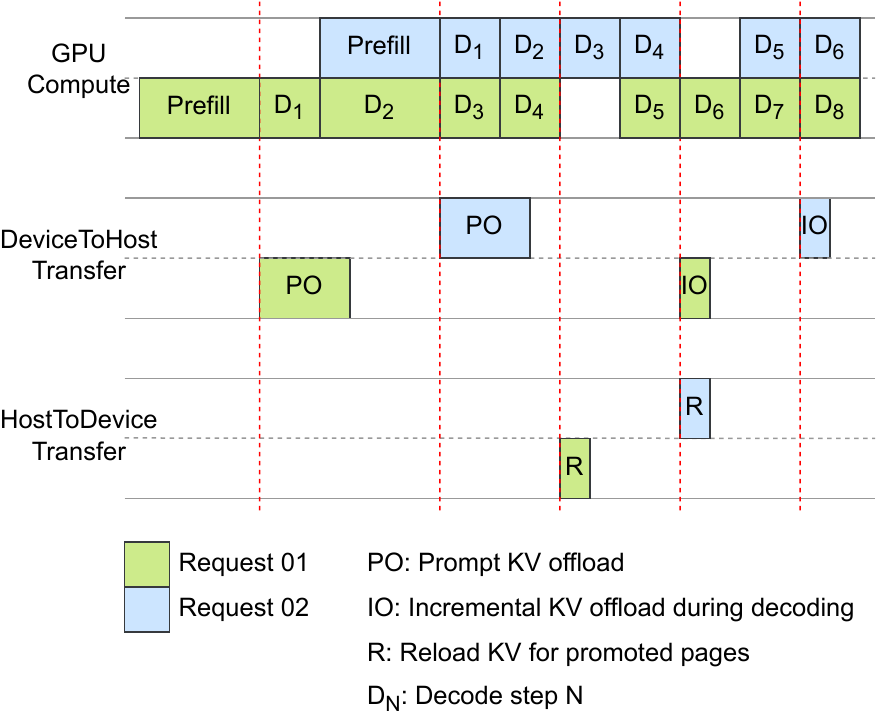}
  \caption{\textbf{FlexiCache Pipeline}. KV offloading is overlapped with the computation of the same request, while KV reloading is overlapped with the computation of other requests in the batch.}
  \label{fig:pipeline-demo}
\end{figure}

\parab{Overlapping Transfer with Compute.}
To prevent KV cache movement from stalling the compute pipeline, transfers between GPU and host memory are carefully overlapped with computation, as shown in Figure~\ref{fig:pipeline-demo}. After the prefill phase, the initial GPU$\rightarrow$host offload begins asynchronously on a low-priority background stream while the main stream continues decoding. Once complete, the block manager releases GPU blocks for pages no longer in the top-K. This significantly reduces per-request GPU memory usage and enables larger batch sizes, as more requests can now fit in GPU memory. Subsequent incremental offloads, triggered as pages become full, are similarly overlapped.

Overlapping host$\rightarrow$GPU transfers are more challenging since promoted KV pages modify the GPU-resident cache. Concurrent decoding of the same request could cause inconsistent reads, particularly when physical block recycling (Section~\ref{sec:physical_block_recycling}) is enabled. To prevent this, the FlexiCache scheduler temporarily pauses only the request being reloaded while other requests continue decoding. The performance impact is minimal: reload sizes are small (Section~\ref{sec:minimize-tx-size}) and typically complete within one decode step. Since different requests pause at different times, only a small fraction of the batch (about 1/16 for a rerank frequency of 16) is idle at any point, keeping overall throughput high.

\parab{Reducing Fragmentation Overhead.}
\label{sec:reduce-fragmentation}
Paged attention scatters the KV cache of each request across the memory pool. A naive GPU$\leftrightarrow$host transfer (e.g., using PyTorch advanced indexing) would first gather the scattered chunks at the source, copy a contiguous buffer, and then scatter them again at the destination. The CPU-side gather/scatter dominates latency in this process.

To eliminate this overhead, we implement a custom CUDA kernel that directly reads from or writes to the pinned host KV cache via Unified Virtual Addressing (UVA), avoiding CPU-side gather/scatter entirely. However, host-mapped reads and writes are limited by PCIe bandwidth, so they occupy GPU Streaming Multiprocessors (SMs) while stalling on I/O. To reduce contention with main compute kernels, we employ three techniques: (1) launching transfers on a low-priority background stream; (2) capping grid size to keep SM occupancy low; and (3) splitting large transfers into multiple smaller chunks, launching a sequence of short kernels. These strategies reduce per-launch stall time and give the CUDA scheduler opportunities to interleave high-priority compute kernels between low-priority transfer kernels.

\subsection{Efficient Block Table Management}
Paged attention maintains a block table that maps each request’s logical indices to physical blocks in GPU memory. In standard dense attention, the KV cache across all heads and layers grows uniformly, allowing a single shared block table for the entire model. FlexiCache breaks this assumption because each head–layer pair independently promotes and evicts KV pages based on its own top-K selection. Hence, their memory layouts diverge, and a single shared block table cannot accurately represent them. FlexiCache therefore performs independent block allocation, deallocation, and mapping for each head–layer pair. To keep this design efficient, we introduce three key optimizations:

\parab{Dirty Tracking.}
Adopting a per-head–layer block mapping scheme expands the block table tensor from $(B, N)$ to $(B, L, H, N)$, where $B$, $L$, $H$, and $N$ denote the batch size, number of layers, number of heads, and maximum number of blocks per sequence, respectively. Typically, this block table is constructed and updated in host memory and then copied to GPU memory at each decoding step. The resulting $L \times H$ increase in table size makes this frequent PCIe transfer a significant performance bottleneck. To mitigate this overhead, FlexiCache tracks which regions of the CPU-side block table have been modified since the last GPU synchronization and transfers only these “dirty” segments. Because the dirty regions are often fragmented, we employ a custom CUDA kernel, similar to the approach in Section~\ref{sec:reduce-fragmentation}, to efficiently transfer the updated segments while avoiding host-side gather overhead.

\parab{Physical Block Reuse During Reranking.}
\label{sec:physical_block_recycling}
During periodic top-K reranking for stable heads, naively deallocating physical blocks of evicted pages and allocating new ones for promoted pages leads to excessive block churn. Frequent calls to the CPU-based block allocator make this process prohibitively expensive. To eliminate this overhead, FlexiCache instead recycles physical blocks. Concretely, we implement a fused CUDA kernel that (i) identifies the sets of evicted and promoted logical blocks, (ii) reassigns physical blocks from the evicted set to the promoted set while updating the GPU-side block table in place, and (iii) generates a mapping from host KV blocks to GPU KV blocks for the subsequent host-to-GPU KV-cache transfer.

\parab{Uniformity Through Null Block.}
Evicting non-critical blocks from GPU memory can create a jagged block table structure, which degrades performance by making simple operations, such as appending new blocks, difficult to vectorize. Moreover, locating a slot in the table by layer, head, and logical index becomes more complex when the layout is irregular. To address this, we enforce a uniform, dense logical structure in the block table using a special null block. When a logical block is deallocated, its physical block is returned to the pool, but its table entry is preserved and redirected to the null block. Because these entries are excluded from attention computation, they are never dereferenced, making the scheme both safe and efficient.

%% file: sections/eval.tex
\section{Evaluation}
\label{evaluation}

\subsection{Setup}
\parab{Implementation.} \rev{FlexiCache is built atop vLLM \cite{kwon2023efficient}, which has key features such as paged KV-cache management and iteration-level scheduling.} We modify vLLM’s Triton-based Flash-Decoding \cite{dao2023flashdecoding} kernel to perform decode attention only on the top-K pages per head. We implement the specialized KV-transfer kernels in CUDA and compile them as PyTorch extensions for seamless integration with vLLM.

\parab{Testbed.} All experiments are conducted on an NVIDIA H100 GPU with 94\,GB of HBM memory. The host system is equipped with 2 AMD~EPYC~9554 64-core processors and 1.1\,TB of DDR5 system memory. Unless otherwise stated, we allocate 180\,GB of host memory for the KV cache. The GPU is connected via PCIe~5.0 with a peak bidirectional bandwidth of 64\,GB/s. Our software environment consists of PyTorch~2.7.0 with CUDA~12.8 and cuDNN~9.5

\parab{Models.} We evaluate FlexiCache on models from two families: Llama-3.1-8B-Instruct \cite{grattafiori2024llama} and Mistral-7B-Instruct-v0.2 \cite{jiang2023clip}.

\parab{Metrics.} For model accuracy, we report the ratio between the benchmark score with FlexiCache and that with dense attention for the same model, providing a uniform measure of performance retention across benchmarks and models despite sparse attention. For system performance, we use total token throughput for offline inference and time per output token (TPOT) for online serving. Although FlexiCache primarily optimizes the decode stage, we also analyze its impact on time-to-first-token (TTFT) in online settings.

\parab{Baselines.} 
\rev{For system performance, we use \textbf{vLLM}’s \cite{kwon2023efficient} Triton-based Flash-Decoding backend as our primary baseline, isolating the impact of FlexiCache’s hierarchical KV placement and stability-aware reranking.}

\rev{For accuracy retention under sparse attention, we compare against \textbf{LServe} \cite{yang2025lserve}, the most conceptually related system. Both frameworks support paged attention and leverage sparsity to reduce the effective token budget for attention computation; however, LServe permanently discards the KV cache of less important tokens, whereas FlexiCache offloads them to host memory for potential reuse. LServe, built atop QServe~\cite{lin2024qserve}, uses quantization by design. To isolate the effect of sparsity on accuracy, we compare its quantized-sparse configuration against its quantized-only baseline and report the relative drop in accuracy. A direct system-level comparison would conflate speedups from quantization and sparsity; hence, we use the vLLM baseline for system performance evaluation.}

\begin{table}[h]
\caption{\rev{\textbf{Cross-task overlap of unstable heads.} Values are the intersection size normalized by the unstable-head set size. GovReport is from LongBench \cite{bai-etal-2024-longbench}; the rest from L-Eval \cite{an-etal-2024-l}.}}
\centering
\footnotesize
\subtable[Mistral-7B-Instruct-v0.2\label{tab:cross-task-overlap-mistral-7b}]{
\centering
\setlength{\tabcolsep}{1pt}
\resizebox{0.98\columnwidth}{!}{%
\begin{tabular}{lcccccccc}
\toprule
\textbf{Dataset} &
\makecell[c]{Open-\\review} &
\makecell[c]{Big-\\Patent} &
\makecell[c]{Multi-\\News} &
\makecell[c]{QM-\\Sum} &
\makecell[c]{Gov-\\Report} &
\makecell[c]{SP-\\ACE} &
\makecell[c]{CU-\\AD} &
\makecell[c]{Summ-\\Screen} \\
\midrule
Openreview   & 1.00 & 0.89 & 0.81 & 0.88 & 0.84 & 0.77 & 0.83 & 0.81 \\
BigPatent    & 0.89 & 1.00 & 0.78 & 0.86 & 0.81 & 0.78 & 0.77 & 0.86 \\
Multi-News   & 0.81 & 0.78 & 1.00 & 0.86 & 0.84 & 0.75 & 0.78 & 0.75 \\
QMSum        & 0.88 & 0.86 & 0.86 & 1.00 & 0.89 & 0.80 & 0.84 & 0.83 \\
GovReport    & 0.84 & 0.81 & 0.84 & 0.89 & 1.00 & 0.77 & 0.80 & 0.80 \\
SPACE        & 0.77 & 0.78 & 0.75 & 0.80 & 0.77 & 1.00 & 0.73 & 0.86 \\
CUAD         & 0.83 & 0.77 & 0.78 & 0.84 & 0.80 & 0.73 & 1.00 & 0.70 \\
SummScreen   & 0.81 & 0.86 & 0.75 & 0.83 & 0.80 & 0.86 & 0.70 & 1.00 \\
\bottomrule
\end{tabular}%
}
}
\subtable[Mistral-Small-24B-Instruct-2501\label{tab:cross-task-overlap-mistral-24b}]{
\centering
\setlength{\tabcolsep}{1pt}
\resizebox{0.98\columnwidth}{!}{%
\begin{tabular}{lcccccccc}
\toprule
\textbf{Dataset} &
\makecell[c]{Open-\\review} &
\makecell[c]{Big-\\Patent} &
\makecell[c]{Multi-\\News} &
\makecell[c]{QM-\\Sum} &
\makecell[c]{Gov-\\Report} &
\makecell[c]{SP-\\ACE} &
\makecell[c]{CU-\\AD} &
\makecell[c]{Summ-\\Screen} \\
\midrule
Openreview   & 1.00 & 0.78 & 0.78 & 0.86 & 0.89 & 0.69 & 0.72 & 0.75 \\
BigPatent    & 0.78 & 1.00 & 0.72 & 0.80 & 0.76 & 0.60 & 0.56 & 0.75 \\
Multi-News   & 0.78 & 0.72 & 1.00 & 0.78 & 0.80 & 0.70 & 0.66 & 0.75 \\
QMSum        & 0.86 & 0.80 & 0.78 & 1.00 & 0.86 & 0.71 & 0.66 & 0.79 \\
GovReport    & 0.89 & 0.76 & 0.80 & 0.86 & 1.00 & 0.69 & 0.68 & 0.79 \\
SPACE        & 0.69 & 0.60 & 0.70 & 0.71 & 0.69 & 1.00 & 0.68 & 0.66 \\
CUAD         & 0.72 & 0.56 & 0.66 & 0.66 & 0.68 & 0.68 & 1.00 & 0.61 \\
SummScreen   & 0.75 & 0.75 & 0.75 & 0.79 & 0.79 & 0.66 & 0.61 & 1.00 \\
\bottomrule
\end{tabular}%
}
}
\subtable[Qwen2.5-32B-Instruct\label{tab:cross-task-overlap-qwen-32b}]{
\centering
\setlength{\tabcolsep}{1pt}
\resizebox{0.98\columnwidth}{!}{%
\begin{tabular}{lcccccccc}
\toprule
\textbf{Dataset} &
\makecell[c]{Open-\\review} &
\makecell[c]{Big-\\Patent} &
\makecell[c]{Multi-\\News} &
\makecell[c]{QM-\\Sum} &
\makecell[c]{Gov-\\Report} &
\makecell[c]{SP-\\ACE} &
\makecell[c]{CU-\\AD} &
\makecell[c]{Summ-\\Screen} \\
\midrule
Openreview   & 1.00 & 0.78 & 0.82 & 0.81 & 0.88 & 0.79 & 0.83 & 0.77 \\
BigPatent    & 0.78 & 1.00 & 0.77 & 0.78 & 0.77 & 0.76 & 0.77 & 0.77 \\
Multi-News   & 0.82 & 0.77 & 1.00 & 0.77 & 0.83 & 0.73 & 0.78 & 0.73 \\
QMSum        & 0.81 & 0.78 & 0.77 & 1.00 & 0.81 & 0.80 & 0.79 & 0.82 \\
GovReport    & 0.88 & 0.77 & 0.83 & 0.81 & 1.00 & 0.80 & 0.79 & 0.77 \\
SPACE        & 0.79 & 0.76 & 0.73 & 0.80 & 0.80 & 1.00 & 0.74 & 0.80 \\
CUAD         & 0.83 & 0.77 & 0.78 & 0.79 & 0.79 & 0.74 & 1.00 & 0.70 \\
SummScreen   & 0.77 & 0.77 & 0.73 & 0.82 & 0.77 & 0.80 & 0.70 & 1.00 \\
\bottomrule
\end{tabular}%
}
}
\label{tab:cross-task-overlap-all}
\end{table}
\rev{\subsection{Model-intrinsic unstable heads}
\label{sec:model-instrinsic-unstable-heads}
Table \ref{tab:cross-task-overlap-all} reports the pairwise cross-task overlap of unstable heads for three models---Mistral-7B-Instruct-v0.2, Mistral-Small-24B-Instruct-2501, and Qwen2.5-32B-Instruct---with mean overlaps of 0.83, 0.76, and 0.81, respectively. These results, along with those in Table \ref{tab:cross-task-overlap}, show that the task-unaware instability of certain KV heads in selecting top KV pages across decode steps generalizes well across models from different families and sizes, consistent with the observations reported in the main paper for Llama. For each model, we classify 25\% of the heads as unstable, corresponding to 64, 80, and 128 heads for Mistral-7B-Instruct-v0.2, Mistral-Small-24BInstruct-2501, and Qwen2.5-32B-Instruct, respectively.}
\subsection{Accuracy}
\parab{Long Context.} We evaluate the accuracy retention of FlexiCache relative to the dense attention baseline across 16 diverse tasks from LongBench \cite{bai-etal-2024-longbench}, as shown in Table~\ref{tab:flexicache-longbench}. The average ratio of FlexiCache to dense attention scores indicates that FlexiCache effectively preserves model performance across both architectures. In this experiment, the top-K page budget is 64 pages (equivalent to 1,024 tokens with a page size of 16), the 64 least stable heads are classified as unstable, and the reranking frequency for stable heads is set to 16. \rev{To prevent task-specific leakage, we compute a single KV-head stability profile on one held-out task and reuse the same stable/unstable head partition for all other tasks. Concretely, we use the KV-head stability profile from the \textit{GovReport} task for all other LongBench tasks, and for \textit{GovReport} itself, we substitute a profile derived from an L-Eval~\cite{an-etal-2024-l} task (\textit{Openreview}).} The strong performance retention further highlights the cross-task stability profile consistency, enabling a single offline profiling step that can be reused at inference time.

\begin{table}[t]
\caption{\textbf{Accuracy retention on LongBench} \cite{bai-etal-2024-longbench}}
\label{tab:flexicache-longbench}
\centering
\setlength{\tabcolsep}{4pt}
\begin{small}
\begin{tabular}{lcccc}
\toprule
& \multicolumn{2}{c}{\textbf{Llama-3.1-8B}} & \multicolumn{2}{c}{\textbf{Mistral-7B-v0.2}} \\
\textbf{Task} & \textbf{Dense} & \textbf{FlexiCache} & \textbf{Dense} & \textbf{FlexiCache} \\
\midrule
NarrativeQA & 29.88 & 29.62 & 20.97 & 20.33 \\
Qasper & 45.41 & 45.73 & 29.41 & 29.49 \\
MultiField-en & 54.81 & 55.20 & 47.16 & 46.91 \\
HotpotQA & 55.97 & 55.06 & 36.99 & 35.39 \\
2WikiMQA & 45.46 & 45.71 & 21.82 & 22.24 \\
Musique & 30.18 & 31.22 & 19.29 & 18.56 \\
GovReport & 34.78 & 34.28 & 33.17 & 31.94 \\
QMSum & 25.27 & 25.01 & 24.10 & 22.94 \\
MultiNews & 27.35 & 27.09 & 27.03 & 26.99 \\
TREC & 72.50 & 72.50 & 71.00 & 71.50 \\
TriviaQA & 91.65 & 91.49 & 85.98 & 86.16 \\
SAMSum & 43.99 & 42.94 & 43.25 & 43.20 \\
PCount & 7.10 & 6.92 & 3.21 & 3.24 \\
PRe & 99.50 & 99.50 & 89.33 & 86.04 \\
Lcc & 54.50 & 54.48 & 49.24 & 49.32 \\
RB-P & 51.65 & 51.88 & 48.61 & 48.90 \\
\midrule
\textbf{Avg. Ratio} & -- & \textbf{1.00} & -- & \textbf{0.99} \\
\bottomrule
\end{tabular}
\end{small}
\end{table}
\begin{table}[h]
\caption{\rev{\textbf{L-Eval task statistics:} Average prompt and generation lengths, and promoted KV size (Llama-3.1-8B, token budget of 2048, 192 stable heads)}}
\label{tab:l-eval-characteristics}
\centering
\begin{small}
\begin{tabular}{cccc}
\toprule
\multicolumn{1}{c}{\textbf{Task}} &
\multicolumn{1}{c}{\shortstack{\textbf{Prompt}\\\textbf{\# Tokens}}} &
\multicolumn{1}{c}{\shortstack{\textbf{Generation}\\\textbf{\# Tokens}}} &
\multicolumn{1}{c}{\shortstack{\textbf{TopK-Delta}\\\textbf{MB}}} \\
\midrule
LongFQA & 5257 & 81 & 46.7 \\
GovReport & 6125 & 377 & 45.0 \\
CUAD & 24906 & 195 & 66.7 \\
QMSum & 15103 & 132 & 66.3 \\
Multi-News & 6002 & 367 & 43.6 \\
Openreview & 10084 & 390 & 55.2 \\
BigPatent & 6363 & 159 & 49.5 \\
SPACE & 17933 & 136 & 67.1 \\
SummScreen & 8882 & 128 & 59.2 \\
\bottomrule
\end{tabular}
\end{small}
\end{table}

\begin{table*}[t]
\caption{\textbf{Accuracy retention on L-Eval} \cite{an-etal-2024-l}}
\label{tab:flexicache-leval}
\centering
\setlength{\tabcolsep}{6pt}
\newcommand{\roth}[1]{\parbox[l]{2mm}{\centering\rotatebox{25}{#1}}} 
\begin{small}
\begin{tabular}{l*{10}{l}}
\toprule
\multicolumn{1}{l}{\parbox[l][2mm][c]{2mm}{\centering Task}} &
\roth{LongFQA} & \roth{GovReport} & \roth{CUAD} &
\roth{QMSum} & \roth{Multi-News} & \roth{Openreview} &
\roth{BigPatent} & \roth{SPACE} & \roth{SummScreen} & \roth{Avg. Ratio} \\
\midrule
\multicolumn{11}{c}{Llama-3.1-8B-Instruct} \\
\midrule
Dense & 49.61 & 27.19 & 37.20 & 17.95 & 17.45 & 21.18 & 33.61 & 16.76 & 15.46 \\
FlexiCache W/O Reranking & 43.28 & 26.74 & 22.63 & 16.85 & 17.27 & 19.22 & 23.12 & 16.40 & 15.58 & 0.89 \\
FlexiCache W/O Unstable Heads & 49.05 & 26.18 & 29.38 & 17.72 & 17.14 & 20.40 & 30.92 & 16.71 & 16.52 & 0.96 \\
FlexiCache 1024 & 51.90 & 28.89 & 37.74 & 17.53 & 17.69 & 20.97 & 31.69 & 16.75 & 13.79 & \textbf{0.99} \\
FlexiCache 2048 & 50.78 & 26.24 & 38.07 & 17.38 & 17.95 & 20.78 & 33.04 & 16.48 & 14.50 & \textbf{0.99} \\
\midrule
\multicolumn{11}{c}{Mistral-7B-Instruct-v0.2} \\
\midrule
Dense & 48.48 & 27.46 & 35.31 & 14.79 & 15.44 & 21.13 & 31.70 & 15.73 & 15.22 \\
FlexiCache W/O Reranking & 41.13 & 24.72 & 28.34 & 14.27 & 15.89 & 18.20 & 22.77 & 14.72 & 13.40 & 0.88 \\
FlexiCache W/O Unstable Heads & 45.96 & 27.69 & 29.18 & 13.63 & 16.43 & 19.78 & 29.06 & 15.84 & 13.48 & 0.95 \\
FlexiCache 1024 & 47.45 & 28.32 & 31.78 & 14.27 & 17.14 & 20.65 & 27.43 & 15.70 & 13.47 & 0.97 \\
FlexiCache 2048 & 46.58 & 28.23 & 34.39 & 14.42 & 16.77 & 20.45 & 30.33 & 16.00 & 13.72 & \textbf{0.99} \\
\midrule
\multicolumn{11}{c}{\rev{Mistral-Small-24B-Instruct-2501}} \\
\midrule
Dense & 48.4 & 24.7 & 33.8 & 16.1 & 16.5 & 21.5 & 28.0 & 15.4 & 14.9 \\
FlexiCache W/O Reranking & 46.7 & 25.5 & 25.9 & 15.5 & 15.9 & 20.5 & 25.9 & 15.2 & 14.2 & 0.944 \\
FlexiCache W/O Unstable Heads & 47.4 & 24.2 & 31.0 & 15.8 & 17.3 & 21.4 & 26.2 & 15.1 & 14.8 & 0.978 \\
FlexiCache 1024 & 48.6 & 25.9 & 29.3 & 16.3 & 16.3 & 21.7 & 26.6 & 15.1 & 14.5 & \textbf{0.981} \\
FlexiCache 2048 & 48.1 & 26.1 & 33.1 & 16.7 & 16.1 & 21.9 & 27.7 & 15.5 & 14.3 & \textbf{1.001} \\
\midrule
\multicolumn{11}{c}{\rev{Qwen2.5-32B-Instruct}} \\
\midrule
Dense & 51.8 & 25.2 & 37.8 & 16.1 & 16.3 & 22.0 & 28.1 & 14.6 & 15.8 \\
FlexiCache W/O Reranking & 49.3 & 23.4 & 28.9 & 15.6 & 15.6 & 21.3 & 25.3 & 13.9 & 16.0 & 0.934 \\
FlexiCache W/O Unstable Heads & 52.1 & 23.0 & 33.6 & 16.2 & 15.4 & 21.6 & 26.0 & 14.5 & 16.6 & 0.967 \\
FlexiCache 1024 & 50.0 & 24.2 & 36.9 & 16.0 & 15.5 & 21.9 & 26.8 & 14.3 & 15.9 & \textbf{0.976} \\
FlexiCache 2048 & 50.9 & 24.7 & 37.7 & 16.0 & 16.2 & 22.0 & 27.9 & 14.4 & 16.6 & \textbf{0.998} \\
\midrule
\multicolumn{11}{c}{Llama-3-8B-Instruct} \\
\midrule
Dense & 45.45 & 34.52 & 33.55 & 15.54 & 16.53 & 18.92 & 31.69 & 15.84 & 14.92 \\
LServe & 45.39 & 32.38 & 28.09 & 14.46 & 16.76 & 18.75 & 29.72 & 16.53 & 11.51 & 0.94 \\
\bottomrule
\end{tabular}
\end{small}
\end{table*}

\parab{Long Context \& Long Generation.}
While LongBench covers a diverse range of tasks, it has a key limitation in evaluating sparse attention: most tasks generate short outputs. Except for a few tasks such as \textit{GovReport} and \textit{MultiNews}, the average generation length is below 50 tokens, with some tasks averaging fewer than 10. As a result, sparse decoding methods that permanently discard the KV cache of less important tokens can still perform well on LongBench because those tokens are rarely needed later in the generation. To address this limitation, we further evaluate FlexiCache on L-Eval~\cite{an-etal-2024-l}, which includes several tasks with long prompts and long generations (as summarized in Table~\ref{tab:l-eval-characteristics}). \rev{To prevent task-specific leakage, we use the KV-head stability profile computed from the \textit{GovReport} task of LongBench.} Unlike permanent-eviction-based sparse decoding, FlexiCache never discards tokens: it retains the full KV cache in host memory and dynamically promotes the newly important pages to GPU memory, preserving model performance during long generation.

\begin{figure*}[t]
  \centering
    \subfigure[Token Throughput for Llama-3.1-8B\label{fig:thrp-llama}]{%
  \includegraphics[width=0.34\textwidth,clip,trim=0 0 273.8535 0]{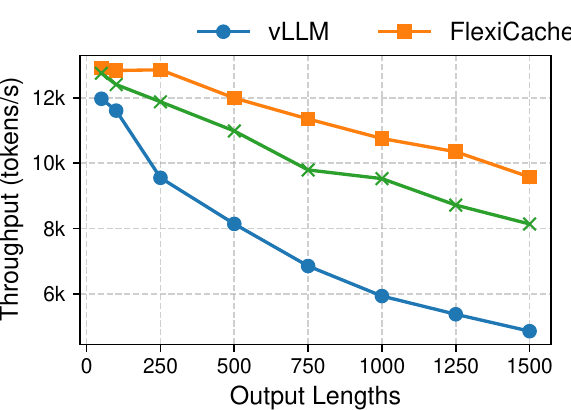}}
  \hspace{-5pt}  
  \subfigure[Token Throughput for Mistral-7B\label{fig:thrp-mistral}]{%
    \includegraphics[width=0.34\textwidth,clip,trim=273.8535 0 0 0]{images/flexicache/throughput_vs_output_length_llama_mistral.pdf}}
  \subfigure[Request Throughput]{
    \includegraphics[width=0.30\textwidth]{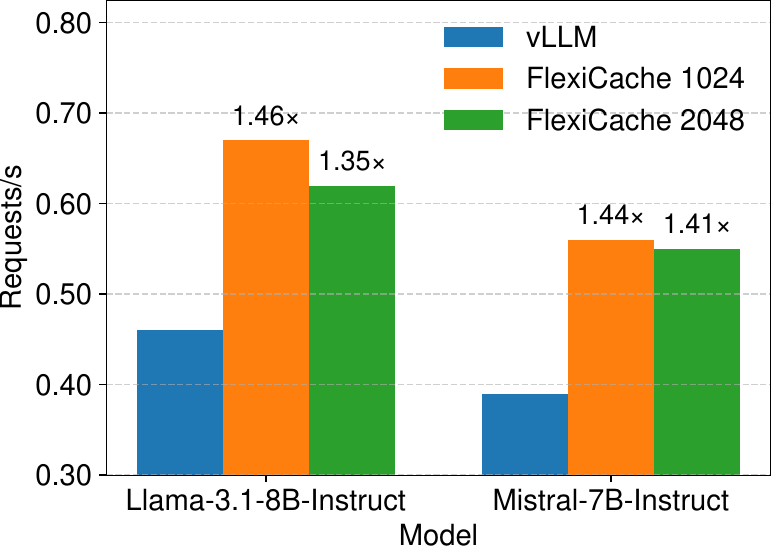}
    \label{fig:request-throughput}
  }
  \caption{\textbf{End-to-end throughput.} FlexiCache consistently outperforms vLLM on both Llama-3.1-8B and Mistral-7B in token throughput, with gains increasing as output length grows. Similar improvements are observed for request throughput with an output length of 500.}
  \label{fig:e2e-throughput}
\end{figure*}

\begin{figure*}[t]
  \centering
    \subfigure[Mean Time Per Output Token\label{fig:mean-tpot}]{%
  \includegraphics[width=0.34\textwidth,clip,trim=0 0 274.584 0]{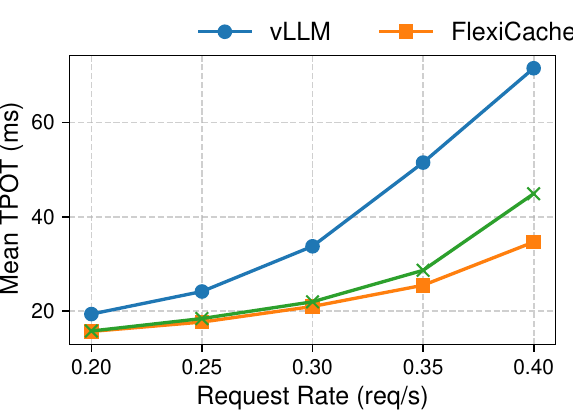}}
  \hspace{-5pt}  
  \subfigure[Mean Time to First Token\label{fig:mean-ttft}]{%
    \includegraphics[width=0.34\textwidth,clip,trim=274.584 0 0 0]{images/flexicache/mean_tpot_ttft_vs_rate.pdf}}
  \subfigure[CDF of Time to First Token\label{fig:cdf-ttft}]{
    \includegraphics[width=0.29\textwidth]{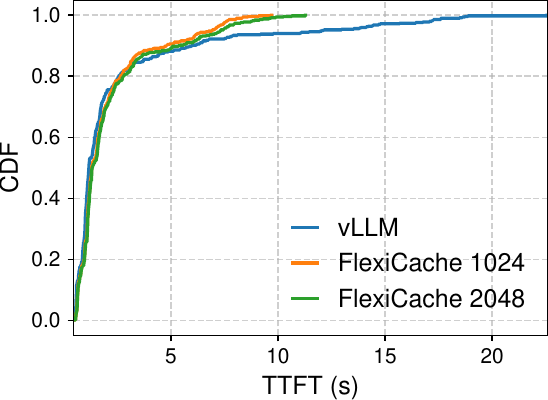}}
  \caption{\textbf{Online serving.} FlexiCache reduces mean TPOT across all arrival rates. By lowering per-request GPU memory usage, it delays TTFT degradation caused by queue buildup at high load, enabling higher sustained request rates while keeping tail TTFT lower.}
  \label{fig:online-serving}
\end{figure*}

Table~\ref{tab:flexicache-leval} compares FlexiCache and LServe~\cite{yang2025lserve} across L-Eval tasks. For FlexiCache, we progressively introduce reranking, stability-aware reranking, and larger token budgets. The final configuration (token budget of 2048, rerank frequency of 16, 64 unstable heads) achieves strong retention of dense-attention accuracy across all models.
For LServe, we follow the default configuration with a token budget of 384 for streaming heads and 4096 for retrieval heads. As LServe is built on top of QServe~\cite{lin2024qserve} and employs a quantized KV cache by design, it cannot be directly compared to a non-quantized dense baseline. To isolate the effect of sparsity, we therefore compare LServe’s quantized-sparse configuration against its quantized-only baseline and report the relative drop resulting from adding sparsity.

\subsection{End-to-end Efficiency}
We evaluate FlexiCache under long-context and long-generation scenarios in both offline and online serving settings. For the workload, we randomly sample prompts from the L-Eval benchmark rather than using synthetic tokens, since the size of the promoted KV cache during reranking depends on workload characteristics. Using real prompts ensures realistic promoted-cache behavior. For Llama-3.1-8B-Instruct with 192 stable heads, fp16 KV cache, and a 2048-token budget, the top-K KV cache for stable heads is 192~MB. The size of the promoted KV varies between 44--67~MB (23--35\%) across tasks (shown in Table~\ref{tab:l-eval-characteristics}). This suggests that temporal stability keeps the promoted set relatively small yet not negligible, underscoring the need for reranking as token importance evolves. All experiments use 64 unstable heads, a stable head rerank frequency of 16, and a token budget of 1024 or 2048.

\begin{figure}[h]
  \centering
  \includegraphics[width=0.8\columnwidth]{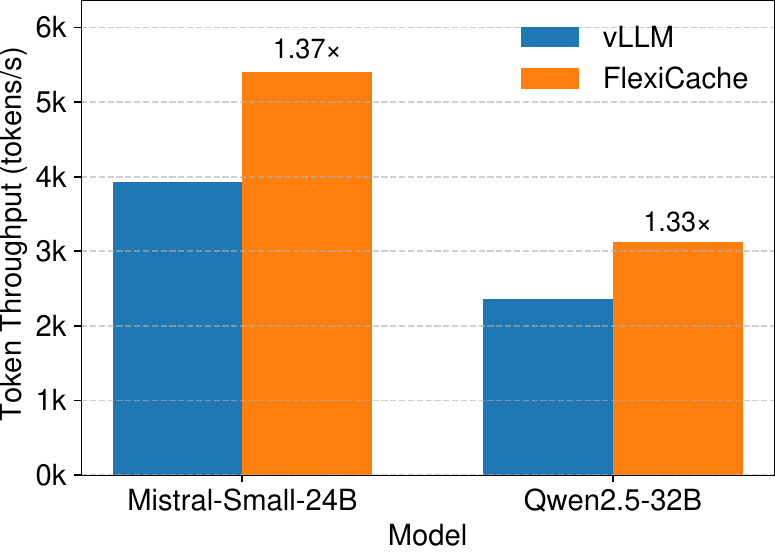}
  \caption{\rev{\textbf{Throughput gain for larger models.} FlexiCache achieves up to 1.37$\times$ improvement in throughput for larger models, with input length randomly selected from 10k--30k tokens and output length set to 500.}}
  \label{fig:throughput-mistral-qwen}
\end{figure}

\parab{Offline Throughput.}
Figure~\ref{fig:e2e-throughput} shows the end-to-end throughput of FlexiCache compared to the vLLM baseline. For both Llama-3.1-8B and Mistral-7B, FlexiCache consistently achieves higher total token throughput (prompt + output tokens per second). Since FlexiCache primarily optimizes the decode phase, the improvement becomes more pronounced as output length increases and decoding dominates the overall latency. The workload comprises 500 randomly sampled requests from L-Eval, with input lengths between 10k–30k and output lengths ranging from 50–1500. The 1024-token budget generally attains higher throughput, with minor fluctuations caused by the randomness of request preemptions. On average across different output lengths, FlexiCache improves token throughput by 1.38–1.55$\times$ for Llama and 1.44–1.46$\times$ for Mistral. On the other hand, as shown in Figure~\ref{fig:request-throughput}, FlexiCache also achieves up to 1.46$\times$ (1024) and 1.41$\times$ (2048) end-to-end request throughput speedups over vLLM for an output length of 500. Figure~\ref{fig:throughput-mistral-qwen} shows that for larger models---MistralSmall-24B-Instruct-2501 and Qwen2.5-32B-Instruct---FlexiCache achieves end-to-end throughput gains of 1.37$\times$ and 1.33$\times$, respectively.

\parab{Online Serving.}
Figure~\ref{fig:online-serving} compares the end-to-end online serving performance of FlexiCache and the vLLM baseline in terms of mean time per output token (TPOT) and time to first token (TTFT). The workload consists of 500 randomly sampled L-Eval requests with input lengths between 10k–30k and output lengths between 20–2000, arriving according to a Poisson process at varying request rates. 

As shown in Figure~\ref{fig:mean-tpot}, FlexiCache consistently outperforms the baseline in mean TPOT, with performance gains becoming more pronounced at higher request rates. At 0.4~req/s, FlexiCache-1024 achieves a mean TPOT of 34.6 compared to 71.5 for the baseline---a 2.1$\times$ improvement---while FlexiCache-2048 reaches 44.9, representing a 1.6$\times$ improvement. Figure~\ref{fig:mean-ttft} shows that TTFT remains similar across systems up to 0.35~req/s, but the baseline collapses beyond this point as GPU memory becomes fully occupied, forcing new requests to wait in the queue and causing TTFT to spike sharply. FlexiCache, by contrast, reduces per-request GPU memory usage during decoding, preventing memory saturation and queue buildup, and thereby maintaining stable TTFT even under higher load. As shown in Figure~\ref{fig:cdf-ttft}, even at moderate load (0.35~req/s), FlexiCache exhibits a shorter tail in TTFT, reflecting fewer queued requests. Overall, FlexiCache sustains higher request rates than the baseline and delivers lower TPOT at any given request rate.

\subsection{Microbenchmarks}
We conduct microbenchmark experiments using the Llama-3.1-8B-Instruct model and FlexiCache configured with a rerank frequency of 16, unless otherwise specified.

\begin{figure}[h]
  \centering
  \includegraphics[width=0.98\columnwidth]{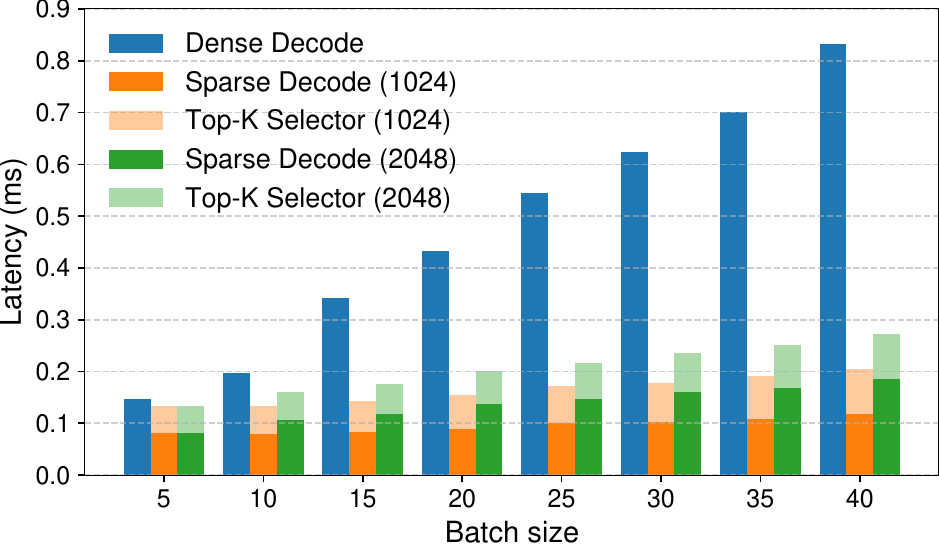}
  \caption{\textbf{Decode Speed vs. Batch Size.} FlexiCache achieves greater speedups as the batch size increases. Each input has 10k tokens.}
  \label{fig:decode-speed-vs-batch-size}
\end{figure}

\parab{Decode Only Speedup.}
Figure~\ref{fig:decode-speed-vs-batch-size} compares the layer-wise decode latency for dense decode and FlexiCache’s sparse decode with token budgets of 1024 and 2048, at different batch sizes with 10k prompt tokens per input. For dense decode, latency reflects the decode kernel execution time, while for FlexiCache, it includes both the top-K selection and sparse decode kernels. \rev{Here, top-K selection measures the time to produce the top-K page indices from the MinMax cache and the current query vector. It runs every step for unstable heads and periodically (e.g., every 16 steps) for stable heads, and we report the time averaged over decoding steps.} As batch size increases, FlexiCache achieves greater speedups (up to 4× at batch size 40 for FlexiCache-1024). At smaller batch sizes, kernel launch overhead dominates, whereas at larger ones, the growing total KV-cache size amplifies the benefit of sparse decoding.

\begin{figure}[h]
  \centering
  \includegraphics[width=0.98\columnwidth]{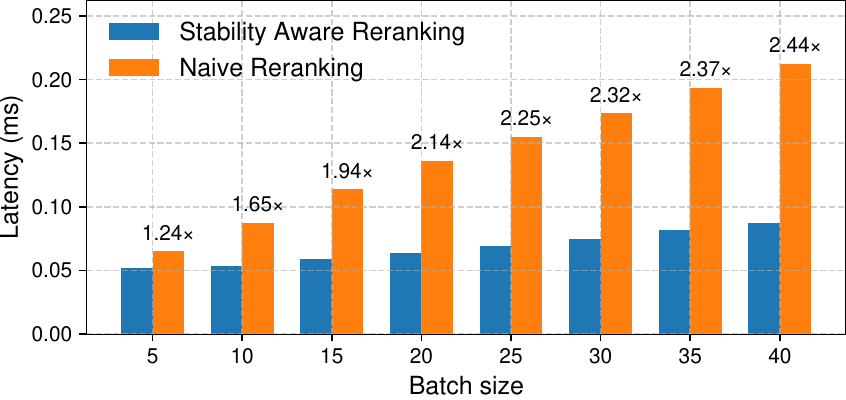}
  \caption{\textbf{Benefit of Stability-Aware Reranking}. Performance gains increase with larger batch sizes. Each input has 10k tokens.}
  \label{fig:stability-aware-reranking}
\end{figure}

\parab{Impact of Stability Aware Reranking.}
Figure~\ref{fig:stability-aware-reranking} shows that stability-aware reranking is 2.44× faster than naive reranking at a batch size of 40, with each input containing 10k tokens. In naive reranking, all heads are reranked at every step. In contrast, stability-aware reranking updates only the 25\% unstable heads at each step and periodically re-ranks the stable ones, keeping the average MinMax cache smaller and reducing latency.

\rev{\parab{Impact of Rerank Frequency and Unstable Head Fraction.}
Table \ref{tab:rerank-unstable} reports the average accuracy retention across L-Eval tasks for Llama-3.1-8B as we vary the unstable head fraction and the rerank interval for stable heads. Accuracy retention improves as we increase the unstable head fraction and decrease the rerank interval. With 25\% unstable heads and a rerank interval of 16, retention remains consistently high. As the rerank interval increases or the unstable head fraction decreases, accuracy degrades gradually rather than catastrophically. We use these same values for other models and retain accuracy without model- or benchmark-specific tuning (Tables \ref{tab:flexicache-longbench} and \ref{tab:rerank-unstable}), suggesting that these hyperparameters act as robust, tunable knobs with graceful trade-offs. Further model- or workload-specific tuning could likely yield additional throughput gains.}

\begin{table}[h]
\caption{\rev{Average accuracy retention on L-Eval across rerank intervals and unstable head fractions.}}
\vspace{2mm}
\label{tab:rerank-unstable}
\centering
\begin{small}
\begin{tabular}{ccccc}
\toprule
& & \multicolumn{3}{c}{\textbf{Unstable Heads Fraction}} \\
\cmidrule(lr){3-5}
& & 0.0625 & 0.125 & 0.25 \\
\midrule
\multirow{2}{*}{\shortstack{\textbf{Rerank}\\\textbf{Interval}}}
& 16 & 0.9692 & 1.0028 & 0.9932 \\
& 32 & 0.9628 & 0.9729 & 0.9936 \\
\bottomrule
\end{tabular}
\end{small}
\end{table}

\begin{figure}[t]
  \centering
  \includegraphics[width=0.9\columnwidth]{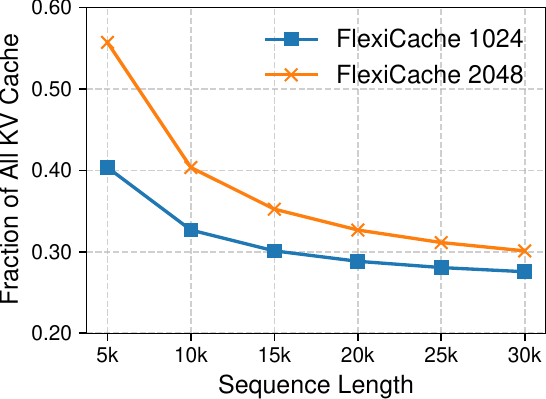}
  \caption{\textbf{GPU Memory Savings.} FlexiCache achieves over 70\% GPU memory savings at sequence lengths $>$20k with a token budget of 1024.
}
  \label{fig:memory-savings}
\end{figure}

\parab{GPU Memory Savings.}
On the GPU, FlexiCache stores the full KV cache for only 25\% of heads and the top-K KV cache for the rest. Figure~\ref{fig:memory-savings} shows the relative GPU memory savings for a single request across sequence lengths with token budgets of 1024 and 2048. As sequence length grows, memory savings asymptotically approach 75\%, reaching about 70\% at 20k tokens and beyond.

%% file: sections/discussion_future.tex
\section{Discussion and Future Work}
\label{discussion-future-work}
\parab{Integration with Other Serving Optimizations.} FlexiCache optimizes the decode phase, with up to 4$\times$ speedup when focusing on the decode kernel (Figure~\ref{fig:decode-speed-vs-batch-size}). We expect larger end-to-end gains when FlexiCache is combined with complementary techniques that optimize other parts of the workflow, such as disaggregating prefill and decode \cite{zhong2024distserve}, kernel fusion, and speculative decoding \cite{leviathan2023fast}.

\parab{Beyond GPU–CPU hierarchy.} FlexiCache currently assumes a two-level GPU–CPU memory hierarchy. This design can be extended to multi-tier hierarchies involving NVMe or distributed memory pools. Efficient KV cache transfer in multi-node environments remains an open challenge, and page-criticality and head-stability–aware KV cache management has the potential to improve existing solutions further~\cite{10.1145/3651890.3672274, cheng2025lmcache}.

\parab{Joint Memory Management and Cluster Scheduling.} In large-scale production serving systems, efficient scheduling and load balancing across GPU clusters remain active challenges. The large KV cache size makes rapid load balancing and task migration difficult \cite{sun2024llumnix}. Temporal stability of heads opens the door for jointly optimizing request batching, KV placement, and prefetch timing using head-level stability signals.

%% file: sections/related.tex
\section{Related Work}
\label{background-and-related-work}

\parab{Sparse Attention.}
Prior work shows that a small subset of critical tokens dominates the attention output at each decode step. Different heuristics are used to identify this subset. These methods fall into two main categories. One class permanently discards less important tokens’ KV entries, such as StreamingLLM \cite{xiao2023efficient} and SnapKV \cite{li2024snapkv}. These methods reduce GPU memory, compute, and I/O but risk accuracy loss in long generations since dropped tokens may later become important. The second class retains the full KV cache but attends only to the most relevant tokens at each step. Quest \cite{10.5555/3692070.3694025} exemplifies this approach by augmenting each KV cache page with per-channel min–max metadata and scoring them against the current query, fetching only the Top-K pages. This reduces compute and I/O while mostly preserving accuracy, though memory use remains unchanged. A hybrid approach, LServe \cite{yang2025lserve}, converts about half of the heads to streaming masks (dropping long-range KV for those heads) and applies dynamic, query-aware selection for the rest. This balances efficiency and accuracy better than pure eviction but still risks accuracy loss in long generations due to permanent KV eviction for half the heads.

\parab{Hierarchical KV Cache Management.} When the KV cache for a batch of requests exceeds GPU memory, a common solution is to offload part of it to host memory or even disk and reload it later. Systems such as FlexGen \cite{10.5555/3618408.3619696} and MoE-Lightning \cite{10.1145/3669940.3707267} adopt this strategy by carefully pipelining data transfers with computation---for example, prefetching the KV cache or weights for the next batch or layer while the current one executes. However, unlike FlexiCache, these designs treat the KV cache within the same layer as uniform; no part is considered more critical for decoding than another. Some of their optimizations, such as layer-wise loading, are orthogonal to FlexiCache and can be used to further augment it. LMCache \cite{cheng2025lmcache} offloads KV cache across the memory hierarchy to reuse common prefix KV across requests and to support prefill–decode disaggregation. This reduces redundant prefill overhead and lowers TTFT, but it does not target offloading the KV cache of requests that are already in the decode phase. NEO \cite{jiang2024neo} uses CPU offloading of KV cache for online serving in order to improve performance over GPU-only systems. However, its offload granularity is sub-batch rather than head-level.

%% file: sections/conclusion.tex
\section{Conclusion}
\label{conclusion}
We observe that KV heads in LLMs exhibit varying degrees of temporal stability in their selection of critical KV pages across decoding steps. Building on this insight, we classify heads into two categories—stable and unstable—and propose FlexiCache, a stability-aware hierarchical KV cache management strategy. FlexiCache dynamically places KV pages across GPU and host memory based on their criticality, with periodic query-aware promotion of previously non-critical pages. This design reduces the GPU’s KV footprint without permanently discarding potentially important tokens—an essential property for long-context, long-generation workloads. 

%% file: sections/artifact_appendix.tex
\appendix
\section{Artifact Appendix}

\subsection{Abstract}

This artifact contains the source code, build scripts, and evaluation scripts needed to reproduce two crucial results for FlexiCache. To run the experiments, an x86\_64 Linux machine with an NVIDIA GPU (80GB minimum), at least 256GB of host memory, and PCIe Gen5 CPU--GPU interconnect is required. All results are based on an NVIDIA H100 GPU (94GB memory). A working CUDA development environment is also required to build FlexiCache, which is implemented on top of vLLM, from source. 

\subsection{Artifact check-list (meta-information)}

{\small
\begin{itemize}[itemsep=0pt, topsep=2pt]
  \item {\bf Algorithm: } Stability-aware hierarchical KV-cache management and sparse decode attention.
  \item {\bf Program: } LongBench evaluation scripts and throughput benchmarking scripts for FlexiCache and vLLM.
  \item {\bf Compilation: } Python package build via \texttt{pip install -e .}, compiling custom CUDA, Triton, and C++ extensions for FlexiCache on top of vLLM.
\item {\bf Model: } meta-llama/Llama-3.1-8B-Instruct, downloaded automatically from Hugging Face by the scripts; requires Internet access and a configured Hugging Face account.
\item {\bf Data set: } zai-org/LongBench, downloaded automatically from Hugging Face when running the scripts; requires Internet access.
  \item {\bf Run-time environment: } Linux, Conda, Python 3.12, PyTorch 2.6.0+cu124, Triton 3.2.0, Transformers 4.50.0, CUDA compatible NVIDIA driver.
  \item {\bf Hardware: } An x86\_64 Linux machine with an NVIDIA GPU (80GB minimum), at least 256GB of host memory, and PCIe Gen5 CPU--GPU interconnect. NVIDIA H100 recommended.
  \item {\bf Run-time state: } For reliable throughput measurements, experiments should be run on an idle machine without other workloads contending for the GPU, CPU, RAM, or host--device bandwidth.
  \item {\bf Execution: } Single-GPU execution via automated bash scripts. The artifact uses a large pinned host-memory pool (about 180GB by default) and frequent host--device KV-cache transfers during execution.
  \item {\bf Metrics: } LLM generation throughput (tokens/second). Long-context benchmarks accuracy.
  \item {\bf Output: } Accuracy numbers and generation throughput presented in 2 CSV files.
  \item {\bf Experiments: } (1) LongBench accuracy retention compared to dense attention baseline (2) end-to-end throughput gain compared to vLLM baseline.
  \item {\bf How much disk space required (approximately)?: } 100GB.
  \item {\bf How much time is needed to prepare workflow (approximately)?: } Around 1 hour to build FlexiCache (implemented on top of vLLM) from source.
  \item {\bf How much time is needed to complete experiments (approximately)?: } Around 2 hours for the accuracy retention experiment. Around 3 hours for the throughput experiment.
  \item {\bf Publicly available?: } Yes.
  \item {\bf Code licenses (if publicly available)?: } Apache License 2.0.
  \item {\bf Data licenses (if publicly available)?: } MIT
  \item {\bf Workflow framework used?: } Bash scripts and Python evaluation code.
  \item {\bf Archived (provide DOI)?: } \href{https://doi.org/10.5281/zenodo.18918856}{https://doi.org/10.5281/zenodo.18918856}
\end{itemize}
}

\subsection{Description}

\subsubsection{How delivered}

The artifact is delivered through the public GitHub repository: \url{https://github.com/NazmulTakbir/FlexiCache}

The repository contains FlexiCache implementation and the evaluation scripts. Reviewers are expected to clone the repository, build the project, and run the provided shell scripts.

\subsubsection{Hardware dependencies}

The artifact is intended to run on a single x86\_64 Linux machine with:
\begin{itemize}[itemsep=0pt, topsep=2pt]
  \item 1$\times$ NVIDIA H100 GPU (94GB recommended, 80GB minimum),
  \item at least 256GB of host DRAM, able to allocate 180GB of pinned host-memory for KV cache offloading
\end{itemize}

\subsubsection{Software dependencies}
\begin{itemize}[nosep]
  \item Linux (we tested on Ubuntu 24.04.2 LTS),
  \item Anaconda (we used 24.9.2),
  \item Python 3.12,
  \item PyTorch 2.6.0+cu124,
  \item Triton 3.2.0,
  \item Transformers 4.50.0,
  \item Datasets 3.6.0,
  \item CUDA NVIDIA driver (we tested on CUDA 12.8).
\end{itemize}
\subsubsection{Data sets}

zai-org/LongBench. Publicly available. Downloaded automatically from Hugging Face when running the scripts; requires Internet access.

\subsection{Installation}

We recommend using Anaconda to isolate the software dependencies. Installation takes approximately 1 hour.

\begin{lstlisting}[style=bashstyle,language=bash]
# 1. Clone the repository
git clone https://github.com/NazmulTakbir/FlexiCache.git
cd FlexiCache
\end{lstlisting}

\begin{lstlisting}[style=bashstyle,language=bash]
# 2. Create and activate a Conda environment
conda create -n FlexiCache python=3.12 -y
conda activate FlexiCache

# 3. set CUDA environment variables so nvcc is found during build (paths may vary by machine)
export CUDA_HOME=/usr/local/cuda-12.8
export CUDACXX=/usr/local/cuda-12.8/bin/nvcc
export PATH=/usr/local/cuda-12.8/bin:$PATH

# 4. Build from source (30min to 1 hour)
export MAX_JOBS=10
pip install -e . 

# 5. Install additional dependencies
pip install -r flexicache_requirements.txt

# 6. Authenticate with Hugging Face (Required for Llama 3.1). Must accept the model agreement on Hugging Face
pip install huggingface_hub
huggingface-cli login
\end{lstlisting}

\subsection{Experiment workflow}

We provide two push-button bash scripts to reproduce the primary accuracy and throughput results for meta-llama/Llama-3.1-8B-Instruct. Both commands should be run from the repository root after installation.

\subsubsection{Accuracy on LongBench}

This script evaluates FlexiCache on a subset of 8 LongBench tasks. This workflow takes approximately 2 hours.

\begin{lstlisting}[style=bashstyle,language=bash]
# Run
bash FlexiCache/LongBench.sh Llama8b

# Results are written to benchmarks/FlexiCache/Language_Modelling/LongBench/longbench_table.csv
\end{lstlisting}

\subsubsection{Throughput}
This script benchmarks end-to-end throughput (tokens/sec) of FlexiCache (with a 1024-token budget) against baseline vLLM for output lengths of 100, 500, 1000, and 1500 tokens, using input prompts randomly sampled between 10k and 30k tokens. This workflow also takes approximately 3 hours.

\begin{lstlisting}[style=bashstyle,language=bash]
# Run
bash FlexiCache/Throughput.sh Llama8b

# Results are written to benchmarks/FlexiCache/Throughput/throughput_table.csv
\end{lstlisting}

\subsection{Evaluation and expected result}

In this section, we provide reference results for comparison. Due to differences in machine environments, absolute throughput values may vary. However, the relative speedup of FlexiCache over the vLLM baseline should remain consistent. Accuracy results may also show small deviations from the reported numbers due to the inherent stochasticity of LLM generation and minor floating-point differences across hardware.

\subsubsection{Accuracy on LongBench}

\begin{table}[h]
\centering
\small
\begin{tabular}{lcc}
\hline
Task & Dense & FlexiCache \\
\hline
Qasper        & 45.23 & 45.92 \\
MultiField-en & 54.93 & 54.94 \\
2WikiMQA      & 45.45 & 45.32 \\
Musique       & 30.18 & 31.22 \\
GovReport     & 34.98 & 34.22 \\
MultiNews     & 27.18 & 27.10 \\
TriviaQA      & 91.64 & 91.49 \\
RB-P          & 49.43 & 49.70 \\
\hline
Avg. Ratio    & --    & 1.00 \\
\hline
\end{tabular}
\caption{Accuracy Retention on LongBench}
\label{tab:ae-longbench}
\end{table}

\subsubsection{Throughput (tokens/second)}

\begin{table}[h]
\centering
\small
\begin{tabular}{lccc}
\hline
Output Length & vLLM & FlexiCache-1024 & Speedup \\
\hline
100  & 11178.7 & 12433.1 & 1.11$\times$ \\
500  & 7742.1  & 11480.2 & 1.48$\times$ \\
1000 & 5609.9  & 9965.4  & 1.78$\times$ \\
1500 & 4549.0  & 9444.9  & 2.08$\times$ \\
\hline
\end{tabular}
\caption{Throughput Gain}
\label{tab:ae-throughput}
\end{table}
\subsection{Experiment customization}

\begin{lstlisting}[style=bashstyle,language=bash]
# Run LongBench & Throughput for other models
bash FlexiCache/LongBench.sh Mistral7b
bash FlexiCache/LongBench.sh Mistral24b
bash FlexiCache/Throughput.sh Mistral7b
bash FlexiCache/Throughput.sh Mistral24b
\end{lstlisting}

\subsection{Methodology}
Submission, reviewing and badging methodology:
\begin{itemize}[nosep]
  \item \url{http://cTuning.org/ae/submission-20190109.html}
  \item \url{http://cTuning.org/ae/reviewing-20190109.html}
  \item \url{https://www.acm.org/publications/policies/artifact-review-badging}
\end{itemize}